# SemAxis: A Lightweight Framework to Characterize Domain-Specific Word Semantics Beyond Sentiment

Jisun An[†]  Haewoon Kwak[†]  Yong-Yeol Ahn[§]

[†]Qatar Computing Research Institute, Hamad Bin Khalifa University, Doha, Qatar
[§]Indiana University, Bloomington, IN, USA

jisun.an@acm.org   haewoon@acm.org   yyahn@iu.edu

## Abstract

Because word semantics can substantially change across communities and contexts, capturing domain-specific word semantics is an important challenge. Here, we propose SEMAXIS, a simple yet powerful framework to characterize word semantics using many semantic axes in word-vector spaces beyond sentiment. We demonstrate that SEMAXIS can capture nuanced semantic representations in multiple online communities. We also show that, when the sentiment axis is examined, SEMAXIS outperforms the state-of-the-art approaches in building domain-specific sentiment lexicons.

## 1 Introduction

In lexicon-based text analysis, a common, tacit assumption is that the meaning of each word does not change significantly across contexts. This approximation, however, falls short because context can strongly alter the meaning of words (Fischer, 1958; Eckert and McConnell-Ginet, 2013; Hovy, 2015; Hamilton et al., 2016b). For instance, the word *kill* may be used much more positively in the context of video games than it would be in a news story; the word *soft* may be used much more negatively in the context of sports than it is in the context of toy animals (Hamilton et al., 2016a). Thus, lexicon-based analysis exhibits a clear limitation when two groups with strongly dissimilar lexical contexts are compared.

Recent breakthroughs in vector-space representation, such as word2vec (Mikolov et al., 2013b), provide new opportunities to tackle this challenge of context-dependence, because in these approaches, the representation of each word is learned from its context. For instance, a recent study shows that a propagation method on the vector space embedding can infer context-dependent sentiment values of words (Hamilton et al., 2016a). Yet, it remains to be seen whether it is possible to generalize this idea to general word semantics other than sentiment.

In this work, we propose SEMAXIS, a lightweight framework to characterize domain-specific word semantics beyond sentiment. SEMAXIS characterizes word semantics with respect to many semantic perspectives in a domain-specific word-vector space. To systematically discover the manifold of word semantics, we induce 732 semantic axes based on the antonym pairs from ConceptNet (Speer et al., 2017). We would like to emphasize that, although some of the induced axes can be considered as an extended version of sentiment analysis, such as an axis of 'respectful' (positive) and 'disrespectful' (negative), some cannot be mapped to a positive and negative relationship, such as 'exogeneous' and 'endogeneous,' and 'loose' and 'monogamous.' Based on this rich set of semantic axes, SEMAXIS captures nuanced semantic representations across corpora. The key contributions of this paper are:

- We propose a general framework to characterize the manifold of domain-specific word semantics.
- We systematically identify semantic axes based on the antonym pairs in ConceptNet.
- We demonstrate that SEMAXIS can capture semantic differences between two corpora.
- We provide a systematic evaluation in comparison to the state-of-the-art, domain-specific sentiment lexicon construction methodologies.

Although the idea of defining a semantic axis and assessing the meaning of a word with a vector projection is not new, it has not been demonstrated that this simple method can effectively in-duce context-aware semantic lexicons. All of the inferred lexicons along with code for SEMAXIS and all methods evaluated are made available in the SEMAXIS package released with this paper[1].

## 2 Related Work

For decades, researchers have been developing computational techniques for text analysis, including: sentiment analysis (Pang and Lee, 2004), stance detection (Biber and Finegan, 1988), point of view (Wiebe, 1994), and opinion mining (Pang and Lee, 2004). The practice of creating and sharing large-scale annotated lexicons also accelerated the research (Stone et al., 1966; Bradley and Lang, 1999; Pennebaker et al., 2001; Dodds et al., 2011; Mohammad et al., 2016, 2017).

These approaches can be roughly grouped into two major categories: lexicon-based approach (Turney, 2002; Taboada et al., 2011) and classification-based approach (Pang et al., 2002; Kim, 2014; Socher et al., 2013). Although the recent advancement of neural networks increases the potential of the latter approach, the former has been widely used for its simplicity and transparency. ANEW (Bradley and Lang, 1999), LIWC (Pennebaker et al., 2001), SO-CAL (Taboada et al., 2011), SentiWordNet (Esuli and Sebastiani, 2006), and LabMT (Dodds et al., 2011) are well-known lexicons.

A clear limitation of the lexicon-based approach is that it overlooks the context-dependent semantic changes. Scholars reported that the meaning of a word can be altered by context, such as communities (Yang and Eisenstein, 2015; Hamilton et al., 2016a), diachronic changes (Hamilton et al., 2016b) or demographic (Rosenthal and McKeown, 2011; Eckert and McConnell-Ginet, 2013; Green, 2002; Hovy, 2015), geography (Trudgill, 1974), political and cultural attitudes (Fischer, 1958) and personal variations (Yang and Eisenstein, 2017). Recently, a few studies have shown the importance of taking such context into accounts. For example, Hovy *et al.* (2015) showed that, by including author demographics such as age and gender, the accuracy of sentiment analysis and topic classification can be improved. It was also shown that, without domain-specific lexicons, the performance of sentiment analysis can be significantly degraded (Hamilton et al., 2016a).

Building domain-specific sentiment lexicons through human input (crowdsourcing or experts) requires not only significant resources but also careful control of biases (Dodds et al., 2011; Mohammad and Turney, 2010). The challenge is exacerbated because 'context' is difficult to concretely operationalize and there can be numerous contexts of interest. For resource-scarce languages, such problems become even more critical (Hong et al., 2013). Automatically building lexicons from web-scale resources (Velikovich et al., 2010; Tang et al., 2014) may solve this problem but poses a severe risk of unintended biases (Loughran and McDonald, 2011).

Inducing domain-specific lexicons from the unlabeled corpora reduces the cost of dictionary building (Hatzivassiloglou and McKeown, 1997; Rothe et al., 2016). Although earlier research utilize syntactic (grammatical) structures (Hatzivassiloglou and McKeown, 1997; Widdows and Dorow, 2002), the approach of learning word-vector representations has gained a lot of momentum (Rothe et al., 2016; Hamilton et al., 2016a; Velikovich et al., 2010; Fast et al., 2016).

The most relevant work is SENTPROP (Hamilton et al., 2016a), which constructs domain-specific sentiment lexicons using graph propagation techniques (Velikovich et al., 2010; Rao and Ravichandran, 2009). In contrast to SENTPROP's sentiment-focused approach, we provide a framework to understand the semantics of words with respect to 732 semantic axes based on ConceptNet (Speer et al., 2017).

## 3 SEMAXIS Framework

Our framework, SEMAXIS, involves three steps: constructing a word embedding, defining a semantic axis, and mapping words onto the semantic axis. Although they seem straightforward, the complexities and challenges in each step can add up. In particular, we tackle the issues of treating small corpora and selecting pole words.

### 3.1 The Basics of SEMAXIS

#### 3.1.1 Building word embeddings

The first step in our approach is to obtain word vectors from a given corpus. In principle, any standard method, such as Positive Pointwise Mutual Information (PPMI), Singular-Value Decomposition (SVD), or `word2vec`, can be used. Here, we use the `word2vec` model because `word2vec` is easier to train and is known to be more robust than competing methods (Levy et al., 2015).

---
[1] https://github.com/ghdi6758/SemAxis

### 3.1.2 Defining a semantic axis and computing the semantic axis vector

A semantic axis is defined by the vector between two sets of 'pole' words that are antonymous to each other. For instance, a sentiment axis can be defined by a set of the most positive words on one end and the most negative words on the other end. Similarly, any antonymous word pair can be used to define a semantic axis (e.g., 'clean' vs. 'dirty' or 'respectful' vs. 'disrespectful').

Once two sets of pole words for the corresponding axis are chosen, we compute the average vector for each pole. Then, by subtracting one vector from the other, we obtain the semantic axis vector that encodes the antonymous relationship between two sets of words. More formally, let $S^+=\{v_1^+, v_2^+, ..., v_n^+\}$ and $S^-=\{v_1^-, v_2^-, ..., v_m^-\}$ be two sets of pole word vectors that have an antonym relationship. Then, the average vectors for each set are computed as $\mathbf{V}^+=\frac{1}{n}\sum_1^n v_i^+$ and $\mathbf{V}^-=\frac{1}{m}\sum_1^m v_j^-$. From the two average vectors, the semantic axis vector, $\mathbf{V}_{\text{axis}}$ (from $S^-$ to $S^+$), can be defined as:

$$\mathbf{V}_{\text{axis}} = \mathbf{V}^+ - \mathbf{V}^- \qquad (1)$$

### 3.1.3 Projecting words onto a semantic axis

Once the semantic axis vector is obtained, we can compute the cosine similarity between the axis vector and a word vector. The resulting cosine similarity captures how closely the word is aligned to the semantic axis. Given a word vector $v_w$ for a word $w$, the score of the word $w$ along with the given semantic axis, $\mathbf{V}_{\text{axis}}$, is computed as:

$$\begin{aligned} score(w)_{\mathbf{V}_{\text{axis}}} &= \cos(v_w, \mathbf{V}_{\text{axis}}) \\ &= \frac{v_w \cdot \mathbf{V}_{\text{axis}}}{\parallel v_w \parallel \parallel \mathbf{V}_{\text{axis}} \parallel} \end{aligned} \qquad (2)$$

A higher score means that the word $w$ is more closely aligned to $S^+$ than $S^-$. When $\mathbf{V}_{\text{axis}}$ is a sentiment axis, a higher score is corresponds to more positive sentiment.

### 3.2 SEMAXIS for Comparative Text Analysis

Although aforementioned steps are conceptually simple, there are two practical challenges: 1) dealing with small corpus and 2) finding good pole words for building a semantic axis.

### 3.2.1 Semantic relations encoded in word embeddings

Since semantic relations are particularly important in our method, we need to ensure that our word embedding maintains general semantic relations. This can be evaluated by analogy tasks. In particular, we use the Google analogy test dataset (Mikolov et al., 2013a), which contains 19,544 questions — 8,869 semantic and 10,675 syntactic questions — in 14 relation types, such as capital-world pairs and opposite relationships.

### 3.2.2 Dealing with small corpus

As in other machine learning tasks, the amount of data critically influences the performance of word embedding methods. However, the corpora of our interest are often too small to facilitate the learning of rich semantic relationships therein. To mitigate this issue, we propose to pre-train a word embedding using a background corpus and update with the target corpora. In doing so, we capture the semantic changes while maintaining general semantic relations offered by the large reference model.

The vector-space embedding drifts from the reference model as we train with the target corpus. If trained too much with the smaller target corpus, it will lose the 'good' initial embedding from the huge reference corpus. If trained too little, it will not be able to capture context-dependent semantic changes. Our goal is thus to minimize the loss in general semantic relations while maximizing the characteristic semantic relations in the new texts.

Consider a corpus of our interest $C$ and a reference corpus $R$. The model $M$ is pre-trained on $R$, and then we start training it on $C$. We use the superscript $e$ to represent the $e$-th epoch of training. That is, $M_C^e$ is the model after the $e$-th epoch trained on $C$. Then, we evaluate the model regarding two aspects: general semantic relations and context-dependent semantic relations. The former is measured by the overall accuracy of the analogy test (Mikolov et al., 2013a). The latter is measured by tracking the semantic changes of the top $k$ words in the given corpus $C$. The semantic changes of the words are measured by the changes in their scores, $\Delta$, on a certain axis; for instance, a sentiment axis, between consecutive epochs. We stop learning when two conditions are satisfied: (1) When the accuracy of the analogy test drops by $\alpha$; and (2) When $\Delta$ is lower than $\beta$. In principle, the model can be updated with the target corpus as long as the accuracy does not drop. We then use $\beta$ to control the epochs. When $\Delta$ is low, the gain by updating the model becomes negligible compared to the cost and thus we can stop updating.

### 3.3 Identifying rich semantic axes for SEMAXIS

The primary advantage of SEMAXIS is that it can be easily extended to examine diverse perspectives of texts as it is lightweight. Although the axis can be defined by any pair of words in principle, we propose a systematic way to define the axes.

#### 3.3.1 732 Pre-defined Semantic Axes

We begin with a pair of antonyms, called *initial pole words*. For instance, a sentiment axis, which is a basis of sentiment analyses, can be defined by a pair of sentiment antonyms, such as 'good' and 'bad.'

To build a comprehensive set of initial pole words, we compile a list of antonyms from ConceptNet 5.5, which is the latest release of a knowledge graph among concepts (Speer et al., 2017). We extract all the antonym concepts marked as '/r/Antonym' edges. Then, we filter out non-English concepts and multi-word concepts. In addition, we eliminate duplicated antonyms that involve synonyms. For instance, only one of the (empower, prohibit) and (empower, forbid) needs to be kept because 'prohibit' and 'forbid' are synonyms, marked as '/r/Synonym' in ConceptNet.

To further refine the antonym pairs, we create a crowdsourcing task on *Figure Eight*, formerly known as *CrowdFlower*. Specifically, we ask crowdworkers: *Do these two words have opposite meanings?* We include those word pairs that a majority of crowdworkers agree to have an opposite meaning. The word pairs that the majority of crowdsource workers disagree were mostly erroneous antonym pairs, such as '5' and '3', and 'have' and 'has.' We then filter out the antonyms that are highly similar to each other. For example, ('advisedly' and 'accidentally') and ('purposely' and 'accidentally') show the cosine similarity of 0.5148, while 'advisedly' and 'purposely' are not marked as synonyms in ConceptNet. Although we use the threshold of 0.4 in this work, a different threshold can be chosen depending on the purpose. Finally, we eliminate concepts that do not appear in the pre-trained Google News 100B word embeddings. As a result, we obtain 732 pairs of antonyms. Each pair of antonyms becomes initial pole words to define one of the diverse axes for SEMAXIS.

We assess the semantic diversity of the axes by computing cosine similarity between every possible pair of the axes. The absolute mean value of the cosine similarity is 0.062, and the standard deviation is 0.050. These low cosine similarity and standard deviation values indicate that the chosen axes have a variety of directions, covering diverse and distinct semantics.

#### 3.3.2 Augmenting pole words

We then expand the two initial pole words to larger sets of pole words, called *expanded pole words*, to obtain more robust results. If we use only two initial pole words to define the corresponding axis, the result will be sensitive to the choice of those words. Since the initial pole words are not necessarily the best combinations possible, we would like to augment it so that it is more robust to the choice of the initial pole words.

To address this issue, we find the $l$ closest words of each initial pole word in the word embedding. We then compute the geometric center (average vector) of $l+1$ words (including the initial pole word) and regard it as the vector representation of that pole of the axis. For instance, refining an axis representing a 'good' and 'bad' relation, we first find the $l$ closest words for each of 'good' and 'bad' and then compute the geometric center of them. The newly computed geometric centers then become both ends of the axis representing a 'good' and 'bad' relation. We demonstrate how this approach improves the explanatory power of an axis describing a corresponding antonym relation in Section 4.3.

## 4 SEMAXIS Validation

In this section, we quantitatively evaluate our approach using the ground-truth data and by comparing our method against the standard baselines and state-of-the-art approaches. We reproduce the evaluation task introduced by Hamilton *et al*. (2016a), recreating Standard English and Twitter sentiment lexicons for evaluation. We then compare the accuracy of sentiment classification with three other methods that generate domain-specific sentiment lexicons.

It is worth noting that we validate SEMAXIS based on a sentiment axis mainly due to the availability of the well-established ground-truth data and evaluation process. Nevertheless, as the sentiment axis in the SEMAXIS framework is not specifically or manually designed but established based on the corresponding pole words, the validation based on the sentiment axis can be generalized to other axes that are similarly established based on other corresponding pole words.

**Standard English:** We use well-known General Inquirer lexicon (Stone et al., 1966) and continuous valence scores collected by Warriner *et al.* (2013) to evaluate the performance of SEMAXIS compared to other state-of-the-art methods. We test all of the methods by using the off-the-shelf Google news embedding constructed from $10^{11}$ tokens (Google, 2013).

**Twitter:** We evaluate our approach with the test dataset from the 2015 SemEval task 10E competition (Rosenthal et al., 2015) using the embedding constructed by Rothe *et al.* (2016).

| Domain | Positive pole words | Negative pole words |
|---|---|---|
| Standard | good, lovely, excellent, fortunate, pleasant, delightful, perfect, loved, love, happy | bad, horrible, poor, unfortunate, unpleasant, disgusting, evil, hated, hate, unhappy |
| Twitter | love, loved, loves, awesome, nice, amazing, best, fantastic, correct, happy | hate, hated, hates, terrible, nasty, awful, worst, horrible, wrong, sad |

Table 1: Manually selected pole words used for the evaluation task in (Hamilton et al., 2016a). These pole words are called seed words in (Hamilton et al., 2016a)

### 4.1 Evaluation Setup

We compare our method against state-of-the-art approaches that generate domain-specific sentiment lexicons.

**State-of-the-art approaches:** Our baseline for the standard English is a WordNet-based method, which performs label propagation over a WordNet-derived graph (San Vicente et al., 2014). For Twitter, we use Sentiment140, a distantly supervised approach that uses signals from emoticons (Mohammad and Turney, 2010). Moreover, on both datasets, we compare against two state-of-the-art sentiment induction methods: DENSIFIER, a method that learns orthogonal transformations of word vectors (Rothe et al., 2016), and SENTPROP, a method with a label propagation approach on word embeddings (Hamilton et al., 2016a). Seed words, which are called pole words in our work, are listed in Table 1.

**Evaluation metrics:** We evaluate the aforementioned approaches according to (i) their binary classification accuracy (positive and negative), (ii) ternary classification performance (positive, neutral, and negative), and (iii) Kendall $\tau$ rank-correlation with continuous human-annotated polarity scores. Since all methods result in sentiment scores of words rather than assigning a class of sentiment, we label words as positive, neutral, or negative using the class-mass normalization method (Zhu et al., 2003). This normalization uses knowledge of the label distribution of a test dataset and simply assigns labels to best match this distribution. For the implementation of other methods, we directly use the source code without any modification or tuning (SocialSent, 2016) used in (Hamilton et al., 2016a).

### 4.2 Evaluation Results

Table 2 summarizes the performance. Surprisingly, SEMAXIS — the simplest approach — outperforms others on both Standard English and Twitter datasets across all measures.

| **Standard English** | | | |
|---|---|---|---|
| Method | AUC | Ternary F1 | Tau |
| SEMAXIS | **92.2** | **61.0** | **0.48** |
| DENSIFIER | 91.0 | 58.2 | 0.46 |
| SENTPROP | 88.4 | 56.1 | 0.41 |
| WordNet | 89.5 | 58.7 | 0.34 |
| **Twitter** | | | |
| Method | AUC | Ternary F1 | Tau |
| SEMAXIS | **90.0** | **59.2** | **0.57** |
| DENSIFIER | 88.5 | 58.8 | 0.55 |
| SENTPROP | 85.0 | 58.2 | 0.50 |
| Sentiment140 | 86.2 | 57.7 | 0.51 |

Table 2: Evaluation results. Our method performs best on both Standard English and Twitter.

### 4.3 Sensitivity to Pole Words

As discussed in Section 3.3.2, because the axes are derived from pole words, the choice of the pole words can significantly affect the performance. We compare the robustness of three methods for selecting pole words: 1) using sentiment lexicons; 2) using two pole words only (initial pole words); and 3) using $l$ closest words on the `word2vec` model as well as the two initial pole words (expanded pole words). For the first, we choose two sets of pole words that have the highest scores and the lowest scores in two widely used sentiment lexicons, ANEW (Bradley and Lang, 1999) and LabMT (Dodds et al., 2011). Then, for the two pole words, we match 1-of-10 positive pole words and 1-of-10 negative pole words in Table 1, resulting in 100 pairs of pole words. For these 100 pairs, in addition to the two initial pole words, we then use the $l$ closest words ($l = 10$) of each of them to evaluate the third method.

We compare these three methods by quantifying how well SEMAXIS performs for the evaluation task. The average AUC for the two pole words method is 78.2. We find that one of the 100 pairs — 'good' and 'bad' — shows the highest AUC (92.4). However, another random pair ('happy' and 'evil') results in the worst performance with the AUC of 67.2. In other words, an axis defined by only two pole words is highly sensitive to the choice of the word pair. By contrast, when an axis is defined by aggregating $l$ closest words, the average AUC increases to 80.6 (the minimum performance is above 71.2). Finally, using pre-established sentiment lexicons results in the worst performance (the AUC of 77.8 for ANEW and 67.5 for LabMT). These results show that identifying an axis is a crucial step in SEMAXIS, and using $l$ closest words in addition to initial pole words is a more robust method to define the axis.

## 5 SEMAXIS in the Wild

We now demonstrate how SEMAXIS can be used in comparative text analysis to capture nuanced linguistic representations beyond the sentiment. As an example, we use Reddit (Reddit, 2005), one of the most popular online communities. Reddit is known to serve diverse sub-communities with different linguistic styles (Zhang et al., 2017). We focus on a few pairs of subreddits that are known to express different views. We also choose them to capture a wide range of topics from politics to religion, entertainment, and daily life to demonstrate the broad applicability of SEMAXIS.

### 5.1 Dataset, Pre-processing, Reference model, and Hyper-parameters

We use Reddit 2016 comment datasets that are publicly available (/u/Dewarim, 2017). We build a corpus from each subreddit by extracting all the comments posted on that subreddit. When the size of two corpora used for comparison is considerably different, we undersample the bigger corpus for a fair comparison. Every corpus then undergoes the same pre-processing, where we first remove punctuation and stop words, then replace URLs with its domain name.

**Reference model for Reddit data** As we discussed earlier, many datasets of our interest are likely too small to obtain good vector representations. For example, two popular subreddits, `/r/The_Donald` and `/r/SandersForPresident`[2], show only 59.8% and 42.1% in analogy test, respectively.[3] Therefore, as we proposed, we first create a pre-trained word embedding with a larger background corpus and perform additional training with target subreddits. We sample 1 million comments from each of the top 200 subreddits, resulting in 20 million comments. Using this sample, we build a word embedding, denoted as Reddit20M, using the CBOW model with a window size of five, a minimum word count of 10, the negative sampling, and down-sampling of frequent terms as suggested in (Levy et al., 2015). For the subsequent training with the target corpora, we train the model with a small starting learning rate, 0.005; Using different rates, such as 0.01 and 0.001, did not make much difference. We further tune the model with the dimension size of 300 and the number of the epoch of 100 using the analogy test results.

| Category | Reddit20M | Google300D |
|---|---|---|
| World | 28.34 | 70.2 |
| family | 94.58 | 90.06 |
| Gram1-9 | 70.21 | 73.40 |
| **Total** | 67.88 | 77.08 |

Table 3: Results of analogy tests, comparing 20M sample texts from Reddit vs. Google 100B News.

Table 3 shows the results of the analogy test using our Reddit20M in comparison with Google300D, which is the Google News embedding used in previous sections. As one can expect, Reddit20M shows worse performance than Google300D. However, the four categories (capital-common-countries, capital-world, currency, and city-in-state denoted by **World**), which require some general knowledge on the world, drive the 10% decrease in overall accu-racy. Other categories show comparable or even better accuracy. For example, Reddit20M outperforms Google300D by 4.52% in the family category. Since Reddit is a US-based online community, the Reddit model may not be able to properly capture semantic relationships in **World** category. By contrast, for the categories for testing grammar (denoted by **Gram1-9**), Reddit20M shows comparable performances with Google300D (70.21 vs. 73.4). In this study, we use Reddit20M as a reference model and update it with new target corpora.

---
[2]`/r/` is a common notation for indicating subreddits.
[3]For both corpora, continuous bag-of-words (CBOW) model with the dimension size of 300 achieves the highest accuracy in the analogy test.

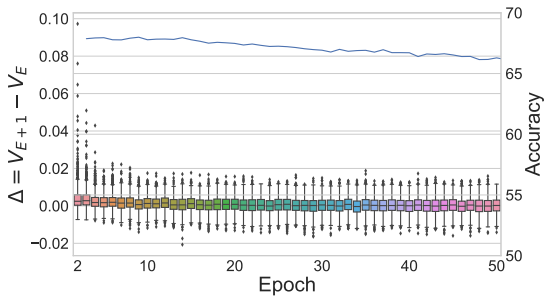

Figure 1: Changes of word semantics (box plot) and accuracy (line graph) over epoch for the model for /r/SandersForPresident

**Updating the reference model** As we explained in Section 3.2.2, we stop updating the reference model depending on the accuracy of the analogy test and semantic changes of the top 1000 words of the given corpus. In our experiments, we set $\alpha = 0.3$ and $\beta = 0.001$. Figure 1 shows the accuracy of the analogy test over epoch as a line plot and the semantic changes of words as a box plot for the model for /r/SandersForPresident. The model gradually loses general semantic relation over epochs, and the characteristic semantic changes stabilize after about 10 epochs. Given the $\alpha$ and $\beta$, we use the embedding after 10 epochs of training with the target subreddit data. We note that the results are consistent when epoch is greater than 10. We choose the number of epoch for other corpora based on the same tactic.

### 5.2 Confirming well-known language representations

Once we have word embeddings for given subreddits by updating the pre-trained model, we can compare the languages of two subreddits. As a case study, we compare supporters of Donald Trump (/r/The_Donald) and Bernie Sanders (/r/SandersForPresident)[4], and examine the semantic differences in diverse issues, such as gun and minority, based on different axes. This can be easily compared with our educated guess learned from the 2016 U.S. Election.

Starting from a topic word (e.g., 'gun') and its closest word (e.g., 'guns'), we compute the average vector of the two words. We then find the closest word from the computed average vector and repeat this process to collect 30 topical terms in each word embedding. Then, we remove words that have appeared less than $n$ times in both corpora. The higher $n$ leads to less coverage of topical terms but eliminate noise. We set $n = 100$ in the following experiments. We consider the remaining words as topic words.

Figure 2 compares how the minority-related terms are depicted in the two subreddits. Figure 2(a) and 2(b) show how minority issues are perceived in two communities with respect to 'Sentiment' and 'Respect' . The x-axis is the value for each word on the Sentiment axis for Trump supporters, and the y-axis is the difference between the value for Trump and Sanders supporters. If the y-value is greater than 0, then it means the word is more 'positive' among Trump supporters compared to that among Sanders supporters.

Some terms perceived more positively (e.g., 'immigration' and 'minorities') while other terms were perceived more negatively ('black', 'latino', 'hispanic') among Trump supporters (Figure 2(a)). As this positive perception on immigration and minorities is unexpected, we examine the actual comments. Through the manual inspection of relevant comments, we find that Trump supporters often mention that they 'agree' with or 'support' the idea of banning immigration, resulting in having a term 'immigration' as more positive than Sanders supporters. However, when examining those words on the 'Disrespect' vs. 'Respect' axis (Figure 2(b)), most of the minority groups are considered disrespectful by Trump supporters compared to Sanders supporters, demonstrating a benefit of examining multiple semantic axes that can reflect rich semantics beyond basic sentiments.

Then, one would expect that 'Gun' would be more positively perceived for Trump supporters compared to for Sanders supporters. Beyond the sentiment, we examine how 'gun' is perceived in

---
[4]Both subreddits have a policy of banning users who post content critical of the candidate. Thus, we assume most of the users in these subreddits are supporters of the candidate.

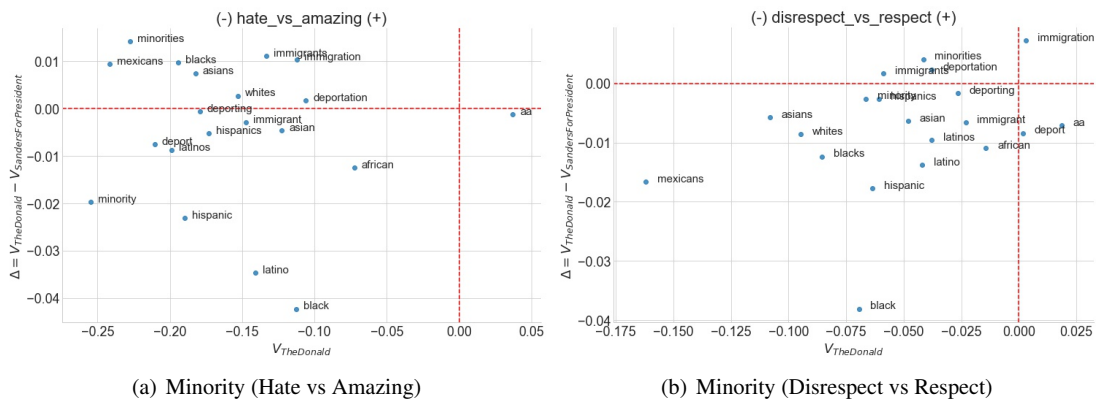

(a) Minority (Hate vs Amazing)   (b) Minority (Disrespect vs Respect)

Figure 2: Trump supporters vs Sanders supporters on Minority issue

two communities for 'Arousal' and 'Safety' axes. We find that Trump supporters are generally positive about gun-related issues, and Sanders supporters associate 'gun' with more arousal and danger.

We also examine two other subreddits: /r/atheism and /r/Christianity. As their names indicate, the former is the subreddit for atheists and the latter is the subreddit for Christians. We expect that the two groups would have different perspectives regarding 'god' and 'evolution.' When examining the four words 'god,' 'pray,' 'evolution,' and 'science' on the 'Unholy' vs. 'Holy' axis, 'god' and 'pray' appear to be more 'holy' in /r/Christianity while 'evolution' and 'science' appear more 'unholy' than in /r/Atheism, which fits in our intuition.

As another example, we examine the /r/PS4 and /r/NintendoSwitch subreddits. PS4 is a video game console released by Sony and Nintendo Switch is released by Nintendo. Although both video game consoles originated from Japan, Nintendo Switch targets more family (children) and casual gamers with more playful and easier games while the games for PS4 target adult and thus tend to be more violent and more difficult to play. We examine three terms ('Nintendo,' 'Mario,' and 'Zelda') from Nintendo Switch and three terms ('Sony,' 'Uncharted,' and 'Killzone') from Sony on the 'Casual' vs. 'Hardcore' axis.[5] We find that 'Mario' and 'Zelda' are perceived more casual in /r/PS4, and 'Uncharted' and 'Killzone' are more hardcore in /r/PS4 than /r/NintendoSwitch. Although both 'Nintendo' and 'Sony' have negative values, 'Nin-tendo' was considered more casual than 'Sony' in /r/PS4. Overall, our method effectively captures context-dependent semantic changes beyond the basic sentiments.

[5]Mario and Zelda are popular Nintendo Switch games, and Uncharted and Killzone are popular PS4 games.

### 5.3 Comparative Text Analysis with Diverse Axes

Let us show how SEMAXIS can find, for a given word, a set of the best axes that describe its semantic. We map the word on our predefined 732 axes, which are explained in Section 3.3.1, and rank the axes based on the projection values on the axes. In other words, the top axes describe the word with the highest strength.

Figure 3(a) shows the top 20 axes with the largest projection values for 'Men' in /r/AskWomen and /r/AskMen, which are the subreddits where people expect replies from each gender. In /r/AskWomen, compared with /r/AskMen, 'Men' seems to be perceived as more vanishing, more established, less systematic, less monogamous, more enthusiastic, less social, more uncluttered, less vulnerable, and more unapologetic. This observed perception of men from women's perspective seems to concur with the common gender stereotype, demonstrating strong potential of SEMAXIS.

Likewise, in Figure 3(b), we examine how a word 'Mario' is perceived in two subreddits /r/NintendoSwitch and /r/PS4. In /r/NintendoSwitch, 'Mario' is perceived, compared with /r/PS4, as more luxurious, famous, unobjectionable, open, capable, likable, successful, loving, honorable, and controllable. On the other hand, users in /r/PS4 consider 'Mario' to be more virtual, creative, durable,

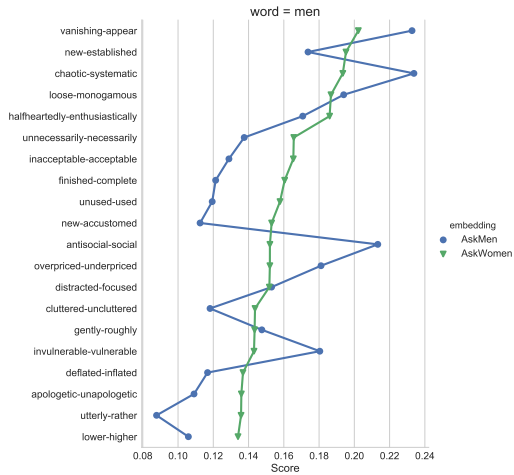
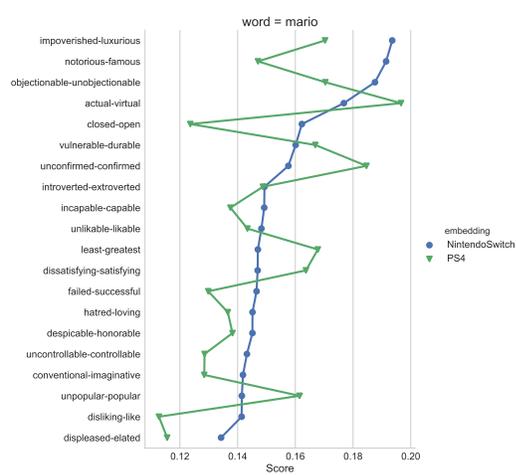

(a) Men in /r/AskWomen and /r/AskMen

(b) Mario in /r/NintendoSwitch and /r/PS4

Figure 3: An example of comparative text analysis using SEMAXIS: (a) 'Men' in `/r/AskWomen` and `/r/AskMen` and (b) 'Mario' in `/r/NintendoSwitch` and `/r/PS4`

satisfying, popular, undetectable, and unstoppable. 'Mario' is perceived more positively in `/r/NintendoSwitch` than in `/r/PS4`, as expected. Furthermore, SEMAXIS reveals detailed and nuanced perceptions of different communities.

## 6 Discussion and Conclusion

We have proposed SEMAXIS to examine a nuanced representation of words based on diverse semantic axes. We have shown that SEMAXIS can construct good domain-specific sentiment lexicons by projecting words on the sentiment axis. We have also demonstrated that our approach can reveal nuanced context-dependence of words through the lens of numerous semantic axes.

There are two major limitations. First, we performed the quantitative evaluation only with the sentiment axis, even though we supplemented it with more qualitative examples. We used the sentiment axis because it is better studied and more methods exist, but ideally it would be better to perform evaluation across many semantic axes. We hope that SEMAXIS can facilitate research on other semantic axes so that we will have labeled datasets for other axes as well. Secondly, Gaffney and Matias (2018) recently reported the Reddit data used in this study is incomplete. The authors suggest using the data with caution, particularly when analyzing user interactions. Although our work examine communities in Reddit, we focus on the difference of the word semantics. Thus, we believe the effect of deleted comment would be marginal in our analyses.

Despite these limitations, we identify the following key implications. First, SEMAXIS offers a framework to examine texts on diverse *semantic axes* beyond the sentiment axis, through the 732 systematically induced semantic axes that capture common antonyms. Our study may facilitate further investigations on context-dependent text analysis techniques and applications. Second, the unsupervised nature of SEMAXIS provides a powerful way to build lexicons of any semantic axis, including the sentiment axis, for non-English languages, particularly the resource-scarce languages.

## 7 Acknowledgements

The authors thank Jaehyuk Park for his helpful comments. This research has been supported in part by Volkswagen Foundation and in part by the Defense Advanced Research Projects Agency (DARPA), W911NF-17-C-0094. The U.S. Government is authorized to reproduce and distribute reprints for Governmental purposes notwithstanding any copyright annotation thereon. The views and conclusions contained herein are those of the authors and should not be interpreted as necessarily representing the official policies or endorsements, either expressed or implied, of DARPA or the U.S. Government.